\documentclass[letterpaper, 10 pt, conference]{ieeeconf}  

\IEEEoverridecommandlockouts                              

\usepackage{cite}
\usepackage{amsmath,amssymb,amsfonts}
\usepackage{algorithmic}
\usepackage{graphicx}
\graphicspath{}
\usepackage{subfigure}
\usepackage{textcomp}
\usepackage{booktabs}
\usepackage{bm}
\usepackage{xcolor}
\usepackage{graphicx}
\usepackage[ruled,vlined]{algorithm2e}
\usepackage{multirow}
\usepackage{soul}
\usepackage{etoolbox}
\makeatletter
\patchcmd{\@makecaption}
  {\scshape}
  {}
  {}
  {}
\makeatother
\usepackage{xcolor}

\title{\LARGE \bf
 Learning-based Hierarchical Control: Emulating the Central Nervous System for Bio-Inspired Legged Robot Locomotion
}

\author{Ge Sun$^{1}$, Milad Shafiee$^{2}$, Peizhuo Li$^{1}$, Guillaume Bellegarda$^{2}$, Auke Ijspeert$^{2}$, and Guillaume Sartoretti$^{1}$ 
\thanks{$^{1}$G. Sun, P. Li, and G. Sartoretti are with the department of Mechanical Engineering, National University of Singapore, 117575 Singapore.
        {\tt\small sunge@u.nus.edu, lipeizhuo@u.nus.edu, mpegas@nus.edu.sg}%
}
\thanks{$^{2}$ M. Shafiee, G. Bellegarda and A. Ijspeert are with the BioRobotics Laboratory, Ecole Polytechnique Federale de Lausanne (EPFL). {\tt\footnotesize (e-mail: firstname.lastname@epfl.ch)}}
}

\begin{document}

\maketitle
\thispagestyle{empty}
\pagestyle{empty}

\begin{abstract}
Animals possess a remarkable ability to navigate challenging terrains, achieved through the interplay of various pathways between the brain, central pattern generators (CPGs) in the spinal cord, and musculoskeletal system. 
Traditional bioinspired control frameworks often rely on a singular control policy that models both higher (supraspinal) and spinal cord functions.
In this work, we build upon our previous research by introducing two distinct neural networks: one tasked with modulating the frequency and amplitude of CPGs to generate the basic locomotor rhythm (referred to as the spinal policy), and the other responsible for receiving environmental perception data and directly modulating the rhythmic output from the spinal policy to execute precise movements on challenging terrains (referred to as the descending modulation policy). 
This division of labor more closely mimics the hierarchical locomotor control systems observed in legged animals, thereby enhancing the robot's ability to navigate various uneven surfaces, including steps, high obstacles, and terrains with gaps. 
Additionally, we investigate the impact of sensorimotor delays within our framework, validating several biological assumptions about animal locomotion systems.
Specifically, we demonstrate that spinal circuits play a crucial role in generating the basic locomotor rhythm, while descending pathways are essential for enabling appropriate gait modifications to accommodate uneven terrain. Notably, our findings also reveal that the multi-layered control inherent in animals exhibits remarkable robustness against time delays.
Through these investigations, this paper contributes to a deeper understanding of the fundamental principles of interplay between spinal and supraspinal mechanisms in biological locomotion. 
It also supports the development of locomotion controllers in parallel to biological structures which are capable of achieving natural locomotion in complex, realistic environments.

\end{abstract}


\section{Introduction}
\label{IROS2024-CPG-modulation-Introduction}
\begin{figure}[t]
	\centering
	\includegraphics[width=1.0\linewidth]{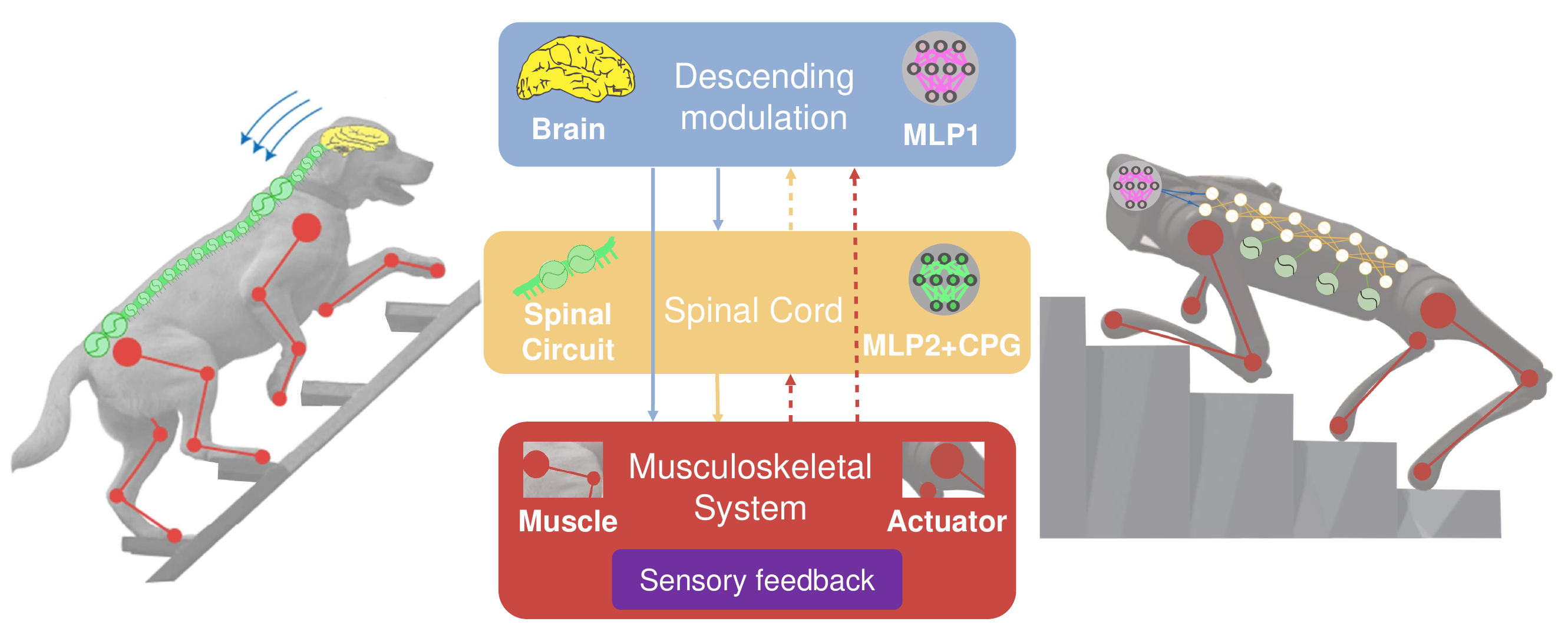}
	\vspace{-0.5cm}
	\caption{Learning-based hierarchical control framework that emulates the mechanisms of the locomotor neural circuits in legged mammals.}
	\label{fig:front_page}
    \vspace{-0.7cm}
\end{figure}

Recent advances in bioinspired robotics have focused on the development of robots that either explore specific biological questions or mimic the mechanisms and designs of biological systems found in nature~\cite{laschi2021bioinspired, melo2023animal,khan2021control}.
This interest is driven by the impressive feats exhibited by many animals, which have evolved sophisticated body forms and cognitive capabilities over millennia.
For instance, legged animals have honed in their abilities and demonstrate extraordinary capabilities to maneuver across a wide variety of terrains~\cite{biewener2018animal}.
This remarkable adaptability depends on complex interactions between many areas of the nervous system, the musculoskeletal system, and the environment.
The agility of animals is even more remarkable given the fact that conduction delays are rather large in neurons~\cite{more2018scaling}, and orders of magnitude higher than in robots' electrical circuits.
At the nervous system level, despite the complex mechanisms at play in animals, locomotion can be primarily attributed to the interplay of three main components: spinal central pattern generators (CPGs), descending modulation, and sensory feedback~\cite{ijspeert2008central}. 
Specifically, CPGs, located in the spinal cord, play a crucial role in animal locomotion as they are responsible for generating coordinated rhythmic muscle movements without rhythmic input.
However, higher brain centers, including the motor cortex, cerebellum, and basal ganglia, refine these rhythmic patterns with descending control signals.
This process enables the adaptation of locomotion to changing environmental conditions, guided by adjustments informed by sensory feedback, such as proprioception and vision.
In this work, we propose a neural structure for locomotion control, comprising two independent networks designed to closely mimic the aforementioned biological structure (Fig.\ref{fig:front_page}).

Studies have shown that the central nervous system hierarchically governs animal locomotion, with control spanning multiple levels, from the cerebral cortex and brainstem down to the spinal cord~\cite{grillner2019current}. 
Firstly, biological evidence highlights the critical role of CPGs located in the spinal cord in providing foundational movement patterns for locomotion, essential for an animal's rapid acquisition of locomotion skills post-birth~\cite{stein1997neurons,vinay2002development}.
Further research demonstrates that self-adaptation can lead to enhancements in spinal circuitry (spinal plasticity), indicating the spinal cord's capacity for learning and memorization~\cite{dietz2003spinal,rygh2002cellular}.
Moreover, in mammals, specific descending pathways, known as the corticospinal tracts, form a crucial link between the cerebral cortex and the spinal cord.
These tracts, unique to mammals, play a vital role in visuomotor coordination, enabling the precise adjustment of movements based on perception feedback to navigate through various terrains and overcome environmental challenges~\cite{ijspeert2002locomotion}.
This innate capacity for rapid learning and adaptation not only sheds light on the fundamental principles of biology, but also fuels considerable interest in the development of robotic locomotion control systems \cite{ijspeert2023integration}.
In particular, we believe that the dynamic interplay between spinal CPGs and supraspinal descending modulation, which is critical for facilitating animal locomotion across diverse terrains, offers a compelling model for controlling robotic locomotion. 

\begin{figure}[t!]
\centering
\includegraphics[width=0.5\textwidth]{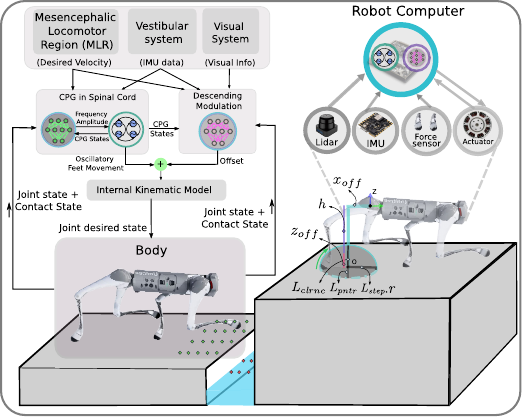}
\caption{\textit{The control diagram of our hierarchical control framework replicates the mechanisms found within the locomotor neural circuits of legged mammals. The spinal network, emulating the spinal cord, generates basic rhythmic gait patterns through CPGs based on internal states. The descending modulation network, representing high-level brain functions, produces signals (offset components) that refine these rhythmic movements in response to internal and environmental information (terrain height map), thereby enabling the robot to navigate complex terrains.}}
 \label{fig:training}
 \vspace{-4mm}
\end{figure}

Drawing inspiration from biology, numerous studies have aimed at replicating the control mechanisms that legged vertebrates use for locomotion~\cite{sharbafi2017bioinspired}.
However, most CPG-based locomotion controllers primarily focus on generating rhythmic motions, which makes it challenging for robots to navigate difficult terrains, which require multiple non-rhythmic adjustments~\cite{liu2011cpg,li2016gait,sun2022joint}.
Additionally, these methods often involve a large number of parameters that require optimization, leading to a time-consuming tuning process.
With the widespread adoption of deep learning in robot controller design, several studies have utilized learning-based methods to fine-tune the parameters and, consequently, the behavior of oscillators, which are used to represent CPGs~\cite{bellegarda2022cpgrl, bellegarda2022visual,deshpande2023deepcpg,wang2021cpg,bellegarda2024quadruped,shafiee2024manyquadrupeds,shafiee2024Viability}.
These adjustments enable improved velocity tracking and omnidirectional locomotion capabilities.
Nevertheless, their ability to traverse challenging environments remains relatively limited.

In this work, we develop a hierarchical control framework that mimics the mechanisms found within the locomotor neural circuits of legged mammals, extending previous work\cite{bellegarda2022cpgrl}. 
Specifically, we employ a learning-based spinal neural network, in conjunction with CPGs, as a fundamental gait pattern generator to represent the spinal cord's function.
We then introduce a supraspinal neural network to modulate the gait patterns generated by the spinal policy, representing the brain's descending pathways, which is integrated with visual information feedback. 
We validate our framework through several experiments in simulation, demonstrating that the robot can adapt to various complex terrains such as stairs, high obstacles, and gaps.
Furthermore, we demonstrate that the spinal policy relies less on sensory feedback to generate rhythmic movements, primarily facilitating locomotion on flat terrain. 
In contrast, the descending modulation policy is more dependent on sensory feedback to modulate gait patterns for challenging terrain, especially for terrains that require discrete and precise movements, such as stairs, high obstacles, and gaps.
Our main contributions can be summarized as follows:
\begin{itemize}
\item{We introduce a learning-based hierarchical structure that independently mimics the functions of the spinal cord and descending modulation mechanisms, utilizing a straightforward and intuitive reward structure.
This bio-inspired framework allows the robot to navigate across challenging terrains, including climbing high platforms, traversing areas with random unevenness and gaps, and ascending stairs that were unachievable with previous neural structures.}
\item {Our framework offers a platform for testing biological hypotheses related to animal locomotion on challenging terrains. In particular, our work explores the unique control mechanisms of rhythmic and discrete movements to enhance the versatility and efficiency of robotic locomotion. 
Furthermore, this framework allows us to investigate the effects of sensorimotor delays on the spinal policy and descending modulation policy for robot behavior on different terrains, shedding light on key biological control principles.}
\item{This work not only deepens our understanding of the function of the central nervous system in biological locomotion, emphasizing the roles of the spinal cord and descending modulation in animal movement strategies, but also facilitates the development of bio-inspired locomotion controllers to enable robots to perform complex and natural movements.}
\end{itemize}


\section{Background}
\label{IROS2024-CPG-modulation-CPG}

Neural networks in the spinal cord, known as CPGs, play a pivotal role in animal locomotion by producing coordinated rhythmic signals, even in the absence of input from the brain and sensory organs~\cite{mackay2002central}.
In this work, we employ non-linear phase oscillators to model the CPG circuits, enabling the generation of rhythmic patterns  similar to those in our previous works~\cite{bellegarda2022cpgrl,ijspeert2007salamander}:
\begin{align}
\ddot{r}_i &= \alpha\left(\frac{\alpha}{4} \left(\mu_i - r_i \right) - \dot{r}_i \right) \label{eq:salamander_r} \\
\dot{\theta}_i &= \nu_i \label{eq:salamander_theta} 
\end{align}
\vspace{-4mm}

\noindent where $r_i$ represents the amplitude of the oscillator, $\theta_i$ denotes the phase of the oscillator, $\nu_i$ and $\mu_i$ are the intrinsic frequency and amplitude, respectively, and $\alpha$ is a factor that represents the convergence rate.

For animals, rhythmic neural activities governed by CPGs are translated into muscle activities, ultimately resulting in rhythmic behaviors~\cite{yu2013survey}.
Building on this principle, we map the outputs of the oscillators to joint positions by first determining the desired positions of the feet in Cartesian space relative to the body frame. Then, we employ inverse kinematics to calculate the desired joint angles. Furthermore, to simulate how signals from higher-level brain regions modulate locomotion behavior in vertebrates, we directly add an offset to the nominal positions of the feet~\cite{shafiee2023puppeteer,shafiee2024Viability}. The formulation of the desired foot positions is as follows:
\begin{align}
x_{i,\text{foot}} &= \ \  -L_{step} (r_i) \cos(\theta_i) + x_{off, i} \label{eq:feet-task1} \\
z_{i,\text{foot}} &= \begin{cases}
    -h+ L_{clrnc}\sin(\theta_i) + z_{off, i} & \text{if } \sin(\theta_i) > 0 \\
    -h+L_{pntr}\sin(\theta_i) + z_{off, i} & \text{otherwise}
\end{cases} 
\label{eq:feet-task2}
\end{align}
\vspace{-4mm}

\noindent where $L_{step}$ represents the nominal step length, $h$ is the nominal body height, $L_{clrnc}$ and $L_{pntr}$ denote the maximum ground clearance during the swing phase and the maximum ground penetration during the stance phase, respectively, and $x_{off}$ and $z_{off}$ directly adjust the desired foot position in the $x$ and $z$ directions, respectively. These parameters are illustrated in Fig.\ref{fig:training}.


\section{LEARNING FRAMEWORK}
\label{IROS2024-CPG-modulation-learningframework}

Our bio-inspired locomotion control framework consists of a spinal policy and a descending modulation policy, both trained through reinforcement learning. 
The spinal policy emulates the spinal cord's function, generating basic gait patterns in conjunction with the CPG, based on internal states and sensory feedback. The descending modulation policy simulates high-level brain functions to produce descending modulation signals~\cite{gratsch2019descending}. These signals fine-tune the movements initiated by the spinal policy, adapting to both internal states and environmental cues, thus enhancing the robot's adaptability in complex settings. 
Fig.\ref{fig:training} depicts the control diagram of our approach, and we detail it further in this section.


\subsection{Spinal Policy}
Biological evidence indicates that the spinal cord’s capacity to perform rhythmic motor patterns, coupled with its access to sensory information specific to activities, enables it to operate with a high degree of automaticity, requiring little to no input from the brain~\cite{fong2009recovery}. 
These findings highlight that the circuits in the spinal cord responsible for locomotion can operate independently of brain control.
Therefore, we employ a spinal policy to represent the learnable circuits in the spinal cord, incorporating the 
CPGs to produce natural cyclic movements. 
To ensure the spinal policy works cooperatively with the descending modulation policy, we enable the spinal policy to access linear velocity commands and proprioceptive information in its observation space. 
The velocity command biologically mimics the signal from the mesencephalic locomotor region (MLR) in the higher brain regions. It is widely acknowledged that the MLR plays an active role in initiating and modulating spinal neural circuitry to control locomotion speed~\cite{ryczko2017nigral}. 
Furthermore, studies show that proprioceptive information, representing the sense of balance and spatial orientation, can be projected to the spinal cord from the vestibular system in vertebrates~\cite{yoo2020neuroanatomy}.
Thus, the spinal policy can dynamically adjust the CPGs' parameters based on the robot's velocity command and proprioceptive feedback from the joint encoders and the Inertial Measurement Unit (IMU), thereby generating a stable and efficient gait.

\subsubsection{Action Space}
To generate basic gait patterns, the spinal policy focuses on adjusting CPG parameters $\bm{a}_{spinal} = \{ \bm{\mu}, \bm{\nu} \}\in \mathbb{R}^{8}$ which primarily control the frequency and phase difference between the legs and the step length (refer to Eq.\ref{eq:salamander_r} and Eq.\ref{eq:salamander_theta}). This design allows the module to not only generate basic gait patterns but also track the commanded velocity, thus enhancing both the adaptability and stability of the locomotion.

\subsubsection{Observation Space}
The observation space includes the operational state of the CPG, denoted as $\bm{x}_{cpg,t} =\{\bm{r}, \dot{\bm{r}}, \bm{\theta}, \dot{\bm{\theta}}\} \in \mathbb{R}^{16}$ (refer to Eq.\ref{eq:salamander_r} and Eq.\ref{eq:salamander_theta}), and previous actions $\bm{a}_{spinal} \in \mathbb{R}^{8}$. Additionally, it encompasses velocity commands $\{v_{x}, v_{y}, \omega \} \in \mathbb{R}^{3}$ in the body frame (x and y directions, as well as yaw rate); contact force booleans $\bm{F}_{GRF} \in \{0,1\}^{4}$; and the current state of the robot $\{\bm{e}_{g}, \bm{v}_{b}, \bm{\omega}_{b} \} \in \mathbb{R}^{9}$, as measured by the Inertial Measurement Unit (IMU), along with the joint states (positions and velocities) represented as $\{\bm{p}_{joint}, \dot{\bm{p}}_{joint}\} \in \mathbb{R}^{24}$. Here, $\bm{e}_{g}$ denotes the projected gravity vector, while $\bm{v}_{b}$ and $\bm{\omega}_{b}$ represent the body's linear and angular velocities, respectively.


\subsection{Descending Modulation Policy}

It should be emphasized that the spinal cord alone is insufficient for locomotion in complex environments. 
The involvement of supraspinal control is crucial, as it provides the necessary drive for movement and the sensorimotor integration required for navigating challenging terrains, as well as performing anticipatory behaviors (as opposed to reactive behaviors)~\cite{norton2010changing}.
There is now accumulating evidence suggesting that supraspinal structures are essential in modulating locomotion, with an emphasis on descending modulation arising from various regions of the supraspinal network~\cite{dubuc2023locomotor}. 
Moreover, these structures are capable of integrating sensory inputs to modify locomotor behavior in response to both internal and external environmental conditions. 
Therefore, we introduce a descending modulation policy to replicate the function by which high-level brain activity modulates rhythmic motion generated by the spinal policy and CPGs.

\subsubsection{Action Space}
In our framework, the descending modulation policy fine-tunes the rhythmic movements produced by the spinal policy and CPGs by introducing specific alterations to the foot's positioning in the horizontal (x) and vertical (z) planes, $\bm{a}_{desc} = \{\bm{x}_{off},\bm{z}_{off}\} \in \mathbb{R}^{8}$ as shown in Eq.\ref{eq:feet-task1} and Eq.\ref{eq:feet-task2}. 
These modifications directly impact the robot's leg movements by altering rhythmic patterns when traversing rough terrain. 
Such adjustments to the rhythmic movement can accommodate uneven terrains, such as stairs. 
Furthermore, it compensates for the limitations of rhythmic gait patterns during locomotion on unstructured terrains, like high obstacles and gaps, which require precise foothold control and discrete movements such as jumping. 
Consequently, the descending modulation policy enhances the robot's adaptability and precision in navigating a variety of terrains.

\subsubsection{Observation Space}

The supraspinal structures involved in controlling locomotion in vertebrates, such as the motor cortex, MLR (Mesencephalic Locomotor Region), and DLR (Diencephalic Locomotor Region), are very complex~\cite{grillner2019current}.
To simplify the descending modulation mechanism, we enable the descending modulation policy to directly access the velocity command and proprioceptive sensory information, which originate from the MLR and the vestibular system, respectively, in vertebrates. 
Additionally, the descending modulation policy can access joint states and ground contact forces through descending spinal tracts, which are the pathways that carry information up and down the spinal cord between the brain and the body. 
Furthermore, by replicating the perceptual capabilities of the higher-level brain regions which process sensory data about the surroundings, we allow the descending modulation policy to access visual information of the environment. 
This is achieved by utilizing a terrain height map, $\bm{H}_{map} \in \mathbb{R}^{17*11}$, which represents a 16 × 10 grid spaced at intervals of 0.05 m, located in front of the robot.
Consequently, the observation space includes both the spinal policy's observations and the terrain height map.
This observation space enables the robot to proactively and adeptly respond to environmental variations. 
Furthermore, it aids in the precise monitoring of each joint's status, ensuring that the robot's movements are responsive and finely tuned to the complexities of the external environment.


\subsection{Training Details}

\begin{figure*}[h!]
\centering
\includegraphics[width=1\textwidth]{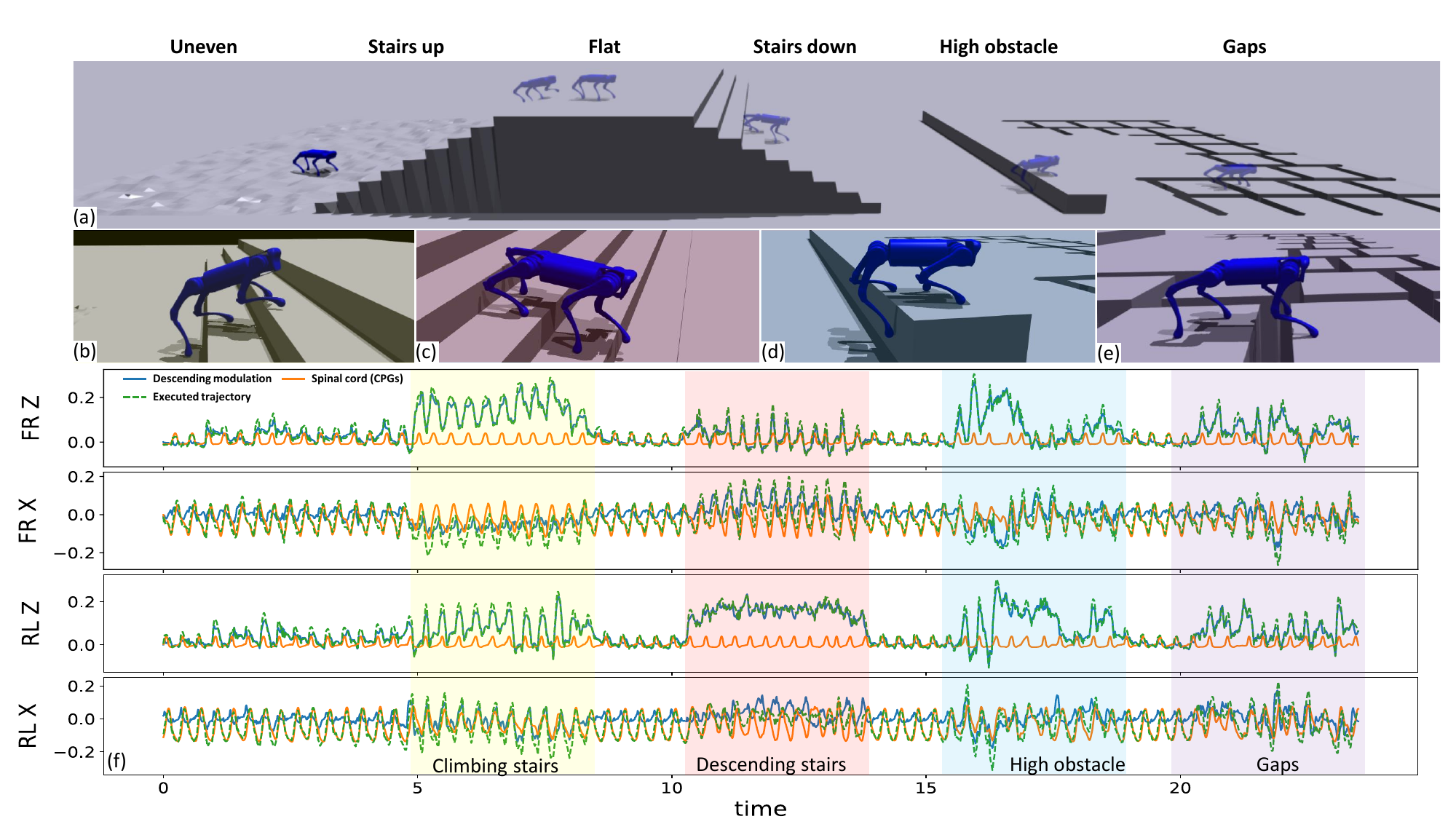}
 \vspace{-4mm} 
\caption{\textit{Rough Terrain Locomotion Experiment: The robot is commanded to move at a speed of 0.9 m/s to travel through the environment.
The duration for traversing each terrain type is indicated by different colors in the plot: yellow for upstairs, red for downstairs, blue for high obstacles, and purple for gaps.
The plot illustrates the output positions from the spinal policy (orange line) and descending modulation policy (blue line), compared to the robot's actual executed position (green dotted line) in both the Z and X dimensions for the front right leg (FR) and the rear left leg (RL). We observe that flat terrain locomotion is mainly governed by the spinal policy, whereas traversal of rough terrains relies mainly on the descending modulation policy.}}
 \label{fig:full_terrain}
 \vspace{-4mm}
\end{figure*}

To train the spinal policy and descending modulation policy, we employ a two-step training process.  
For the first training phase, the robot is trained on flat terrain with the spinal policy only. 
As shown in Table~\ref{shaping_rewards_table}, we reward velocity tracking and locomotion distance, and we penalize the robot's orientation (to ensure stability), as well as power to promote energy efficiency.
For the second training phase, we train the descending modulation policy in conjunction with the spinal policy, whose parameters are frozen, on complex terrains.
To ensure continuity between the two training phases, we make only minor adjustments to the reward function, keeping  the reward structures for the two policies as similar as possible, while allowing the robot to explore and adapt to challenging environments.  Specifically, we reduce the penalty for foot contact forces from 0.1$dt$ to 0.01$dt$ for rough terrain locomotion. Additionally, we introduce a penalty for actions generated by the descending modulation policy to minimize unnecessary impacts on basic gait patterns, and another penalty term that penalizes foot contact with vertical surfaces.
The second phase focuses on various uneven terrains, including steps (up to 18 cm in height for each step), high platforms (up to 50 cm high), randomly varying terrains (with heights within a range of [-10cm, 10cm]), and gaps (up to 50 cm in width), as illustrated in Fig.\ref{fig:full_terrain}.
Velocity commands $v_{x}^{cmd} $ are randomly sampled from 0.5 m/s to 2 m/s.
Both policies operate at a control frequency of 100 Hz, while the torques, derived from desired joint positions, are updated at a frequency of 1 kHz.

We simulate the Unitree quadruped robot A1 in Isaac Gym, which utilizes PhysX as the physics engine, on a single NVIDIA RTX 4090 GPU~\cite{rudin2022learning}.
We employ Proximal Policy Optimization (PPO) to train both policies~\cite{schulman2017proximal} using the same hyperparameters as in our previous work~\cite{bellegarda2022cpgrl}.
The spinal policy and the descending modulation policy are implemented using a 3-layer Multi-Layer Perceptron (MLP) with hidden layers of dimensions [512, 256, 128].

\begin{table}[tbp]
\caption{\textit{Reward function for the two training phases: the reward terms for the first phase (spinal policy) are detailed at the top, and the second training phase (descending modulation policy) follows the same structure as the first, with additional reward terms introduced at the bottom.   $\phi{(x)}$ denotes the squared exponential $e^{-||x||^2/\sigma}$, where $\sigma=0.25$. $\Phi$ represents the orientation information of the base link, $\bm{P}_{base}$ indicates the position of the base, and $F$ represents the foot contact force, with $F_{\text{max}} = 180$.}}
	\centering
 \resizebox{\linewidth}{!}{
	\begin{tabular}{ c c c }
		\hline
		\textbf{Reward term } 		& \textbf{Expression} & $\textbf{w}$ \\
		\hline
        $linear \ velocity$	& $\phi(\bm{v}^{cmd}_t - \bm{v}_t)$	& $10dt$   \\
        $orientation$		& $-(||\Phi_{pitch}||^2 + ||\Phi_{roll}||^2)$	& $20dt$  \\
		$orientation_{yaw}$		& $-||\Phi_{yaw}||^2$ & $30dt$	\\
		$power$		& $-||\bm{\tau}_t \cdot \bm{\dot{q}}_t||$	& $0.01dt$\\
		$locomotion \ distance$		& $\text{min}(\bm{P}_{base,t} - \bm{P}_{base,t-1}, \bm{v}^{cmd}_t \cdot dt) $	& $800dt$ \\
		$feet \ contact \ forces$		& $-\sum_i \max(\|{\bm{F}}_i\| - F_{\text{max}}, 0)$	& $0.1dt (0.01dt)$ \\
        \hline
		$feet \ stumble$	& $-\text{Any}\left( \sqrt{F_{x}^2+F_{y}^2} > 5 \times |F_z| \right)
$	& $1dt$\\
  		$action$		& $-||\bm{a}_{desc}||^2$	& $0.001dt$ \\
		\hline
	\end{tabular}}
    \vspace{-3mm}
\label{shaping_rewards_table}
\end{table}

\section{Experiments and discussion}
\label{IROS2024-CPG-modulation-Results}
In this section, we present simulation results of deploying our learned policies on a quadruped robot to traverse various terrains. These experiments explore the interaction between the spinal cord policy and the descending modulation policy across different terrains, while also verifying certain biological assumptions. Furthermore, we examine the impact of sensorimotor delay on these two policies during locomotion. A supplementary video offers clear visualizations of the discussed experiments.

\begin{figure}[t]
	\centering
\includegraphics[width=0.48\textwidth]{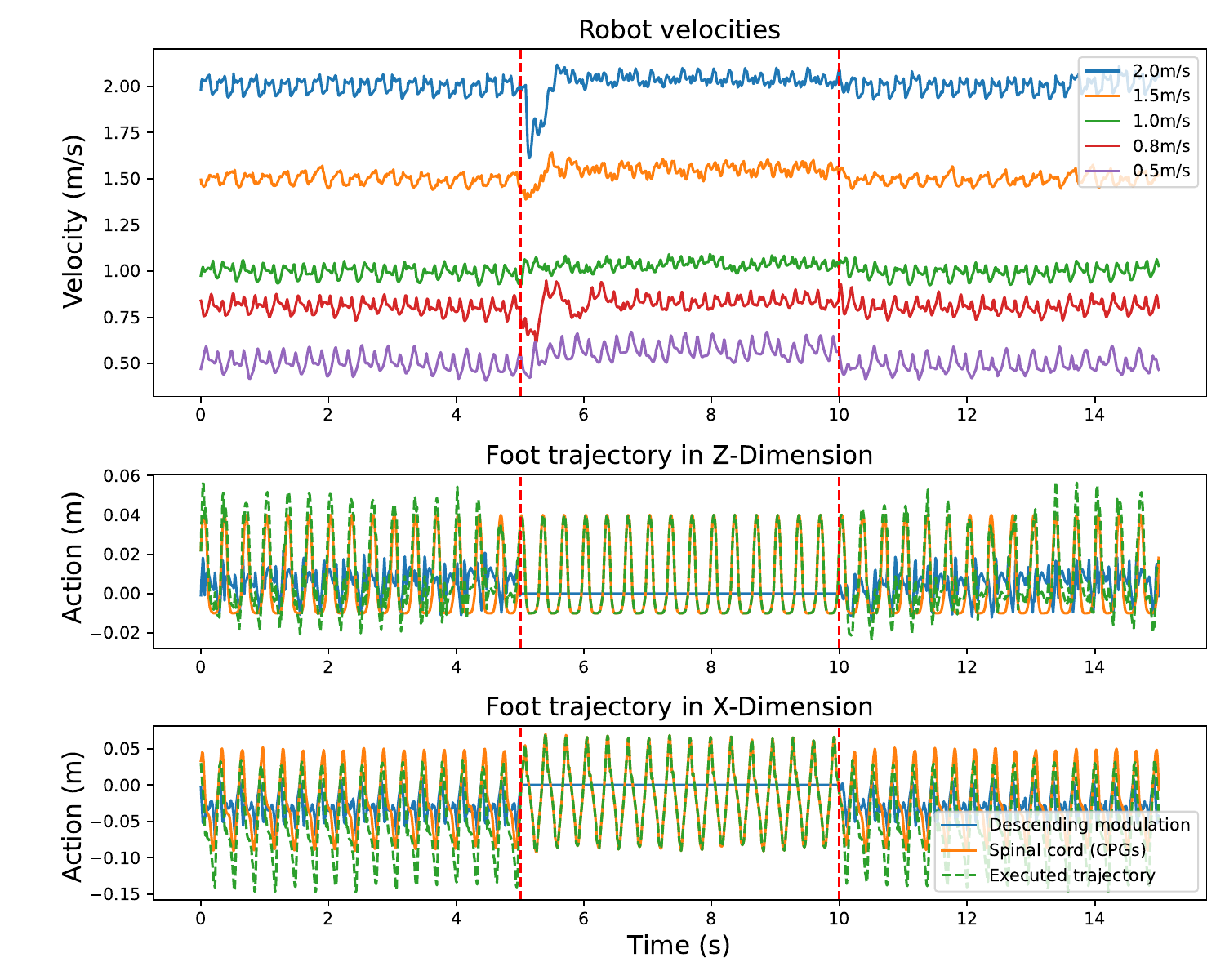}
\vspace{-4mm}
	\caption{\textit{Flat Terrain Locomotion Experiment: Initially, the robot's movement on flat terrain is governed by both the spinal policy and the descending modulation policy. The descending modulation policy is deactivated at t=5s and reactivated at t=10s. The top figure illustrates the robot's tracking velocity across varying commanded velocities. Meanwhile, the middle section compares the output positions from the spinal policy and the descending modulation policy to the robot's actual executed position (front right leg) in the Z dimension, with a commanded speed of 1.0 m/s. The bottom section presents this comparison in the X dimension. This indicates that the robot's movement on flat terrain is primarily controlled by the spinal policy.} }
	\label{fig:flat_terrain}	
	\vspace{-4.0mm} 
\end{figure}

\subsection{Flat Terrain Locomotion}
We first investigate the roles of the spinal and descending modulation policies in regulating locomotion across flat terrain.
Initially, the robot is commanded to walk forward at speeds ranging from 0.5 m/s to 2 m/s on flat terrain. While the spinal policy remains activated, the descending modulation policy is deactivated at $t = 5s$ and then re-activated at $t=10s$. 
We observe that neither the robot's velocity nor its behavior are significantly altered, as demonstrated in Fig.\ref{fig:flat_terrain}, indicating that for the basic locomotion (movement on flat terrain), the behavior is primarily governed by the spinal cord neural network.
The top figure shows that, with only the spinal policy active, the robot can still track the commanded velocity through quick self-adjustment. The bottom plots show the outputs for the front right leg from the spinal policy and descending modulation policy, alongside the executed position of the front right leg. We observe that the executed trajectory (indicated by the green dotted line) largely overlaps with the output of the spinal neural network (represented by the orange line). This indicates that the descending modulation policy exerts minimal influence on locomotion across flat terrain, supporting the biological perspective that basic locomotion patterns are governed by the spinal cord~\cite{hultborn2007spinal}.

\begin{figure*}[t!]
\centering
\includegraphics[width=1\textwidth]{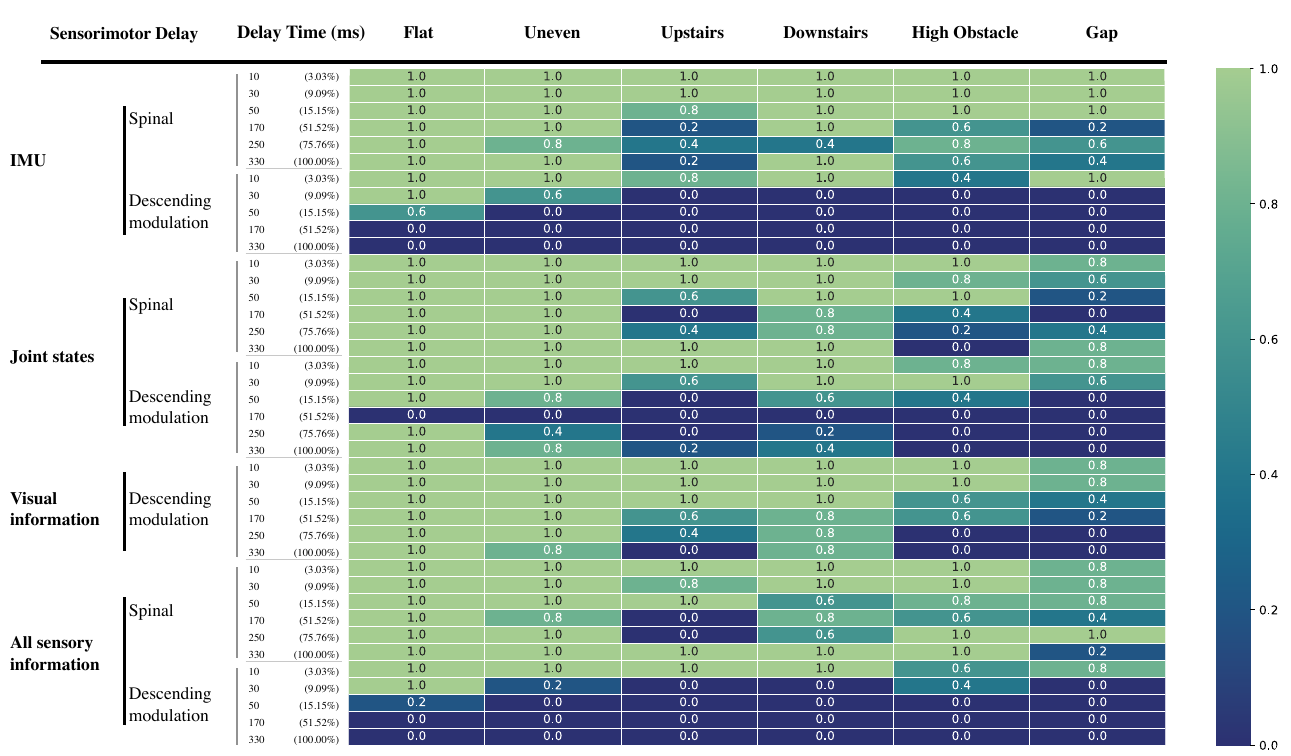}
 \vspace{-4mm} 
\caption{\textit{Success rates for sensorimotor delay experiments: The robot is commanded to travel at 0.9 m/s across six types of terrain—flat, uneven, upstairs, downstairs, high obstacles, and gaps. Each terrain tests the impact of varying sensory delays on two control policies: the spinal policy and the descending modulation policy. Each experiment involves five trials, activating both policies simultaneously, with the robot initiating each trial from distinct starting positions (for example, a success rate of 0.6 indicates three successes and two failures). It's important to note that both policies tested were trained without sensorimotor delays. Additionally, the figure displays the delay time and its corresponding percentage of a single gait cycle.}}
 \label{fig:sensory_delay}
 \vspace{-4mm}
\end{figure*}

\subsection{Rough Terrain Locomotion}
We further test the framework across four different types of rough terrains: uneven surfaces (randomly generated elevations ranging in [-0.08, 0.08] meters), stairs (with a step width of 31 cm and step height of 18 cm), high obstacles (up to 30 cm), and gaps (with lengths up to 25 cm), which are similar but randomized compared with the training environment, as illustrated in Fig.\ref{fig:full_terrain}(a).
In comparison, the dimensions of the Unitree A1, when in the nominal standing configuration, are 50 cm x 30 cm x 40 cm~\cite{unitreeA1}. 
Starting from different initial positions on flat terrain with commanded speeds ranging from 0.5 m/s to 2 m/s, the robot successfully traverses all of these terrains without a single fall in 10 trials.
To examine the contributions of the spinal policy and descending modulation policy to locomotion on rough terrain, we plot the outputs of both neural networks alongside the executed positions of the front right leg and rear left leg in Fig.\ref{fig:full_terrain}(f). 
This analysis reveals that, on rough terrain, the spinal policy (indicated by the orange line) continues to generate rhythmic outputs that represent the basic gait patterns, while the descending modulation policy (indicated by the blue line) significantly modulates the movements produced by the spinal policy. 
Our observations indicate that, during rough terrain locomotion, the step height and footfall are primarily managed by the descending modulation policy, as evidenced by the outputs in the $z$ and $x$ dimensions, respectively.

To facilitate ascending up stairs, the movements of the front legs are modulated by the descending modulation policy to draw them nearer to the body, thus lowering the forebody  (see the plot of the front right leg in the $z$ dimension around $t = 5s$). 
Similarly, to realize descending down stairs, the movements of the rear legs are modulated by the descending modulation policy to bring them closer to the body (see the plot of the rear left leg in z dimension around $t = 12s$). 
Interestingly, a dynamic jumping behavior is observed when crossing a high obstacle of the same height as the robot, with the descending modulation policy modulating the movement by sending an impulse signal to the rear legs (see the plot of the rear left leg in the $z$ dimension around $t = 17s$). 
For gap crossing, the descending modulation policy contributes not only to the foot trajectory in the $z$ dimension to lower the body height for stable crossing, but also to the foot trajectory in the $x$ dimension to control the foothold placement to avoid stepping into the gaps (see the plots of both legs in the $z$ and $x$ dimensions around $t = 23s$). 
These experimental results corroborate the biological view that the basic locomotor rhythm is centrally generated by spinal circuits, and that descending pathways are crucial for ensuring appropriate modifications of gait to accommodate uneven terrain~\cite{drew2004cortical}.

\subsection{Sensorimotor Delay}
Whether an animal is engaging in dynamic gait locomotion in complex environments or moving on flat terrain, the effectiveness of its sensory feedback is constrained by sensorimotor delays, which are non-negligible in animals~\cite{more2018scaling}.
To investigate the influence of sensorimotor delay on the spinal policy and descending modulation policy, we assess the robot's performance across six types of terrain: flat, uneven, upstairs, downstairs, high obstacles, and gaps. 
We assess the robot's performance across terrains similar to those in the previous experiment but introduce varying amounts of sensorimotor delays.
Sensory delays vary in duration, with delay times ranging from 10 ms to 330 ms across different experiments. 
The robot's commanded speed is consistently set at 0.9 m/s. 
It is important to note that both policies are trained without sensorimotor delay and are activated throughout all experiments.
Each experiment consists of 5 trials, with the robot starting from different initial positions on each terrain. Successfully traversing the entire specified terrain is considered a success, and we record the success rate for each experiment. 
Sensory information is categorized into three groups: IMU (vestibular sensory information), joint states (proprioceptive sensory information), and visual sensory information (terrain height map, analogous to visual sensory information in animals). 
Additionally, we test the impact of delays across all sensory groups.
The success rate for each experiment is depicted as a heat map in Fig.\ref{fig:sensory_delay}.

Firstly, we find that the descending modulation policy is particularly sensitive to IMU sensory delays, especially on complex terrains such as stairs, high obstacles, and gaps. With a delay time of 30 ms, the robot falls while traversing these terrains. In contrast, the spinal policy exhibits relative robustness to IMU sensory delays. The robot still has the capability to successfully travel through all the terrains with IMU sensory delays up to 330 ms, which is equal to one gait cycle time. Furthermore, IMU sensory delays in the spinal policy have less effect on traversing simple terrains like flat and uneven terrains.

Secondly, we observe that delays in sensing joint states adversely affects the robot, particularly when navigating stairs, over high obstacles, and across gaps. Such terrains require precise movements in order to successfully traverse them. Interestingly, we observe that the performance with a delay time equal to one gait cycle (330 ms) is better than that with half a gait cycle (170 ms) for joint state sensory delays. We believe that the one gait cycle delay time makes the delayed joint state sensory feedback mostly coincide with the current joint states because of the rhythmic movement. 
Moreover, the spinal policy is more sensitive to delays in joint states than to IMU sensory feedback.
Since the spinal policy lacks access to visual information, visual sensory delays are applied exclusively to the descending modulation policy. Surprisingly, visual sensory delays have a more minor impact on crossing stairs, especially when descending, which can still be managed with a specific rhythmic locomotion pattern. However, terrains requiring discrete motions, such as jumping over high obstacles and precise foothold placement for gaps, are significantly affected by visual sensory delays. For all types of sensory delays, we find that the descending modulation policy is more adversely affected than the spinal policy. With all sensory feedback delayed, the descending modulation policy can even impact the gait on flat terrain.

Overall, we conclude that, as could be expected, for flat terrain locomotion, the spinal policy relies less on sensory feedback to generate rhythmic movements, as its performance on flat terrain is nearly unaffected by sensory delays. In contrast, the descending modulation policy is more sensitive to sensorimotor delays, indicating a greater reliance on sensory feedback to modulate gait patterns. This is particularly true for terrains requiring discrete and precise movements, such as climbing stairs, navigating over high obstacles, and traversing gaps.
Our experimental results align with the biological perspective that sensory feedback is integrated at multiple levels, and the relative contributions of feedforward (as provided by the CPG, and to some extent by the descending policy) and feedback (as provided by the spinal and descending policies) control can vary depending on the terrain and the risks of falling\cite{ijspeert2023integration}. Specifically, the spinal cord can produce rhythmic output even when movement-related afferent input is eliminated \cite{dietz2003spinal}, and sensory signals contribute to shaping locomotor output and adapting it to environmental demands through descending pathways\cite{saradjian2015sensory}.

\section{Conclusion}
\label{IROS2024-CPG-modulation-Conclusion}
In this work, we propose a hierarchical, bio-inspired control framework for legged robot locomotion that mimics the central nervous system of animals. 
This framework integrates a spinal neural network tasked with adapting the key parameters of a conventional CPG model, emulating the spinal cord's role in generating basic rhythmic locomotion patterns.
Additionally, our framework includes a descending modulation policy that learns to directly modulate this rhythmic movement, representing the descending pathway between the brain and spinal cord.
We verified the trained framework's performance across various rough terrains and examined the contributions of the two policies to locomotion on these terrains. Additionally, we investigated the effects of sensorimotor delay on both policies. 
Our experimental results indicate that the spinal policy is less sensitive to sensory feedback when generating rhythmic movements for locomotion on flat terrain, whereas the descending modulation policy relies more on sensory feedback, especially for terrains requiring discrete and precise movements, such as climbing stairs, navigating over high obstacles, and traversing gaps. 
Our control architecture proves quite robust against sensorimotor delays, even without being trained with any delays. 
We speculate that training with delays (fixed or random) could potentially enhance this robustness, a hypothesis warranting future investigation. Furthermore, mechanical properties can also aid in handling sensorimotor delays, as demonstrated in~\cite{ashtiani2021hybrid}. 
Overall, the proposed framework not only demonstrates effectiveness on a robotic platform, but also validates several biological assumptions concerning animal locomotion systems.

Future work will focus on investigating the inner signal passing between the spinal policy and the descending modulation policy by concurrently training both policies online, thereby aiming to replicate the intricate connections between the spinal cord and supraspinal structures. Specifically, we will delve into how descending modulation can infer changes and adjust the central pattern generator (CPG), such as its frequency, amplitude, or even gait, and modulate reflexes by altering feedback gains, reflecting mechanisms observed in animals. Furthermore, we plan to apply our control framework to a hardware platform, enabling us to comprehensively assess its reliability and performance.



\bibliographystyle{IEEEtran}
\bibliography{CPG_modulation} 

\begin{thebibliography}{10}
\providecommand{\url}[1]{#1}
\csname url@rmstyle\endcsname
\providecommand{\newblock}{\relax}
\providecommand{\bibinfo}[2]{#2}
\providecommand\BIBentrySTDinterwordspacing{\spaceskip=0pt\relax}
\providecommand\BIBentryALTinterwordstretchfactor{4}
\providecommand\BIBentryALTinterwordspacing{\spaceskip=\fontdimen2\font plus
\BIBentryALTinterwordstretchfactor\fontdimen3\font minus
  \fontdimen4\font\relax}
\providecommand\BIBforeignlanguage[2]{{%
\expandafter\ifx\csname l@#1\endcsname\relax
\typeout{** WARNING: IEEEtran.bst: No hyphenation pattern has been}%
\typeout{** loaded for the language `#1'. Using the pattern for}%
\typeout{** the default language instead.}%
\else
\language=\csname l@#1\endcsname
\fi
#2}}

\bibitem{laschi2021bioinspired}
C.~Laschi and B.~Mazzolai, ``Bioinspired materials and approaches for soft
  robotics,'' \emph{Mrs Bulletin}, vol.~46, pp. 345--349, 2021.

\bibitem{melo2023animal}
K.~Melo, T.~Horvat, and A.~J. Ijspeert, ``Animal robots in the african
  wilderness: Lessons learned and outlook for field robotics,'' \emph{Science
  Robotics}, vol.~8, no.~85, p. eadd8662, 2023.

\bibitem{khan2021control}
A.~T. Khan, S.~Li, and X.~Cao, ``Control framework for cooperative robots in
  smart home using bio-inspired neural network,'' \emph{Measurement}, vol. 167,
  p. 108253, 2021.

\bibitem{biewener2018animal}
A.~Biewener and S.~Patek, \emph{Animal locomotion}.\hskip 1em plus 0.5em minus
  0.4em\relax Oxford University Press, 2018.

\bibitem{more2018scaling}
H.~L. More and J.~M. Donelan, ``Scaling of sensorimotor delays in terrestrial
  mammals,'' \emph{Proceedings of the Royal Society B}, vol. 285, no. 1885, p.
  20180613, 2018.

\bibitem{ijspeert2008central}
A.~J. Ijspeert, ``Central pattern generators for locomotion control in animals
  and robots: a review,'' \emph{Neural networks}, vol.~21, no.~4, pp. 642--653,
  2008.

\bibitem{grillner2019current}
S.~Grillner and A.~El~Manira, ``Current principles of motor control, with
  special reference to vertebrate locomotion,'' \emph{Physiological reviews},
  2019.

\bibitem{stein1997neurons}
P.~S. Stein, \emph{Neurons, networks, and motor behavior}.\hskip 1em plus 0.5em
  minus 0.4em\relax MIT press, 1997.

\bibitem{vinay2002development}
L.~Vinay, F.~Brocard, F.~Clarac, J.-C. Norreel, E.~Pearlstein, and J.-F.
  Pflieger, ``Development of posture and locomotion: an interplay of
  endogenously generated activities and neurotrophic actions by descending
  pathways,'' \emph{Brain Research Reviews}, vol.~40, no. 1-3, pp. 118--129,
  2002.

\bibitem{dietz2003spinal}
V.~Dietz, ``Spinal cord pattern generators for locomotion,'' \emph{Clinical
  Neurophysiology}, vol. 114, no.~8, pp. 1379--1389, 2003.

\bibitem{rygh2002cellular}
L.~J. Rygh, A.~Tj{\o}lsen, K.~Hole, and F.~Svendsen, ``Cellular memory in
  spinal nociceptive circuitry.'' \emph{Scandinavian journal of psychology},
  vol.~43, no.~2, pp. 153--159, 2002.

\bibitem{ijspeert2002locomotion}
A.~Ijspeert, ``Locomotion, vertebrate,'' \emph{The handbook of brain theory and
  neural networks}, pp. 649--654, 2002.

\bibitem{ijspeert2023integration}
A.~J. Ijspeert and M.~A. Daley, ``Integration of feedforward and feedback
  control in the neuromechanics of vertebrate locomotion: a review of
  experimental, simulation and robotic studies,'' \emph{Journal of Experimental
  Biology}, vol. 226, no.~15, p. jeb245784, 2023.

\bibitem{sharbafi2017bioinspired}
M.~A. Sharbafi and A.~Seyfarth, \emph{Bioinspired legged locomotion: models,
  concepts, control and applications}.\hskip 1em plus 0.5em minus 0.4em\relax
  Butterworth-Heinemann, 2017.

\bibitem{liu2011cpg}
C.~Liu, Q.~Chen, and D.~Wang, ``Cpg-inspired workspace trajectory generation
  and adaptive locomotion control for quadruped robots,'' \emph{IEEE
  Transactions on Systems, Man, and Cybernetics, Part B (Cybernetics)},
  vol.~41, no.~3, pp. 867--880, 2011.

\bibitem{li2016gait}
J.~Li, J.~Wang, S.~X. Yang, K.~Zhou, H.~Tang, \emph{et~al.}, ``Gait planning
  and stability control of a quadruped robot,'' \emph{Computational
  intelligence and neuroscience}, vol. 2016, 2016.

\bibitem{sun2022joint}
G.~Sun and G.~Sartoretti, ``Joint-space cpg for safe foothold planning and body
  pose control during locomotion and climbing,'' \emph{IEEE Robotics and
  Automation Letters}, vol.~7, no.~4, pp. 9889--9896, 2022.

\bibitem{bellegarda2022cpgrl}
G.~Bellegarda and A.~Ijspeert, ``{CPG-RL}: Learning central pattern generators
  for quadruped locomotion,'' \emph{IEEE Robotics and Automation Letters},
  vol.~7, no.~4, pp. 12\,547--12\,554, 2022.

\bibitem{bellegarda2022visual}
G.~Bellegarda, M.~Shafiee, and A.~Ijspeert, ``Visual {CPG-RL}: Learning central
  pattern generators for visually-guided quadruped locomotion,'' in \emph{2024
  IEEE International Conference on Robotics and Automation (ICRA)}, 2024.

\bibitem{deshpande2023deepcpg}
A.~M. Deshpande, E.~Hurd, A.~A. Minai, and M.~Kumar, ``Deepcpg policies for
  robot locomotion,'' \emph{IEEE Transactions on Cognitive and Developmental
  Systems}, 2023.

\bibitem{wang2021cpg}
J.~Wang, C.~Hu, and Y.~Zhu, ``Cpg-based hierarchical locomotion control for
  modular quadrupedal robots using deep reinforcement learning,'' \emph{IEEE
  Robotics and Automation Letters}, vol.~6, no.~4, pp. 7193--7200, 2021.

\bibitem{bellegarda2024quadruped}
G.~Bellegarda, M.~Shafiee, M.~E. {\"O}zberk, and A.~Ijspeert, ``Quadruped-frog:
  Rapid online optimization of continuous quadruped jumping,'' \emph{arXiv
  preprint arXiv:2403.06954}, 2024.

\bibitem{shafiee2024manyquadrupeds}
M.~Shafiee, G.~Bellegarda, and A.~Ijspeert, ``Manyquadrupeds: Learning a single
  locomotion policy for diverse quadruped robots,'' \emph{2024 IEEE
  International Conference on Robotics and Automation}, 2024.

\bibitem{shafiee2024Viability}
M.~Shafiee, G.~Bellegarda, and A.~Ijspeert, ``Viability leads to the emergence
  of gait transitions in learning agile quadrupedal locomotion on challenging
  terrains,'' \emph{Nature Communications}, vol.~15, no.~1, p. 3073, 2024.

\bibitem{mackay2002central}
M.~MacKay-Lyons, ``Central pattern generation of locomotion: a review of the
  evidence,'' \emph{Physical therapy}, vol.~82, no.~1, pp. 69--83, 2002.

\bibitem{ijspeert2007salamander}
A.~J. Ijspeert, A.~Crespi, D.~Ryczko, and J.-M. Cabelguen, ``From swimming to
  walking with a salamander robot driven by a spinal cord model,''
  \emph{Science}, vol. 315, no. 5817, pp. 1416--1420, 2007.

\bibitem{yu2013survey}
J.~Yu, M.~Tan, J.~Chen, and J.~Zhang, ``A survey on cpg-inspired control models
  and system implementation,'' \emph{IEEE transactions on neural networks and
  learning systems}, vol.~25, no.~3, pp. 441--456, 2013.

\bibitem{shafiee2023puppeteer}
M.~Shafiee, G.~Bellegarda, and A.~Ijspeert, ``Puppeteer and marionette:
  Learning anticipatory quadrupedal locomotion based on interactions of a
  central pattern generator and supraspinal drive,'' \emph{2023 IEEE
  International Conference on Robotics and Automation}, 2023.

\bibitem{gratsch2019descending}
S.~Gr{\"a}tsch, A.~B{\"u}schges, and R.~Dubuc, ``Descending control of
  locomotor circuits,'' \emph{Current Opinion in Physiology}, vol.~8, pp.
  94--98, 2019.

\bibitem{fong2009recovery}
A.~J. Fong, R.~R. Roy, R.~M. Ichiyama, I.~Lavrov, G.~Courtine, Y.~Gerasimenko,
  Y.~Tai, J.~Burdick, and V.~R. Edgerton, ``Recovery of control of posture and
  locomotion after a spinal cord injury: solutions staring us in the face,''
  \emph{Progress in brain research}, vol. 175, pp. 393--418, 2009.

\bibitem{ryczko2017nigral}
D.~Ryczko, S.~Gr{\"a}tsch, L.~Schl{\"a}ger, A.~Keuyalian, Z.~Boukhatem,
  C.~Garcia, F.~Auclair, A.~B{\"u}schges, and R.~Dubuc, ``Nigral glutamatergic
  neurons control the speed of locomotion,'' \emph{Journal of Neuroscience},
  vol.~37, no.~40, pp. 9759--9770, 2017.

\bibitem{yoo2020neuroanatomy}
H.~Yoo and D.~M. Mihaila, ``Neuroanatomy, vestibular pathways,'' 2020.

\bibitem{norton2010changing}
J.~Norton, ``Changing our thinking about walking,'' \emph{The Journal of
  physiology}, vol. 588, no. Pt 22, p. 4341, 2010.

\bibitem{dubuc2023locomotor}
R.~Dubuc, J.-M. Cabelguen, and D.~Ryczko, ``Locomotor pattern generation and
  descending control: a historical perspective,'' \emph{Journal of
  Neurophysiology}, vol. 130, no.~2, pp. 401--416, 2023.

\bibitem{rudin2022learning}
N.~Rudin, D.~Hoeller, P.~Reist, and M.~Hutter, ``Learning to walk in minutes
  using massively parallel deep reinforcement learning,'' in \emph{Conference
  on Robot Learning}.\hskip 1em plus 0.5em minus 0.4em\relax PMLR, 2022, pp.
  91--100.

\bibitem{schulman2017proximal}
J.~Schulman, F.~Wolski, P.~Dhariwal, A.~Radford, and O.~Klimov, ``Proximal
  policy optimization algorithms,'' \emph{arXiv preprint arXiv:1707.06347},
  2017.

\bibitem{hultborn2007spinal}
H.~Hultborn and J.~B. Nielsen, ``Spinal control of locomotion--from cat to
  man,'' \emph{Acta Physiologica}, vol. 189, no.~2, pp. 111--121, 2007.

\bibitem{unitreeA1}
U.~Robotics, ``Unitree a1 - high performance quadruped robot,''
  \url{https://unitreerobotics.net/robotdog/unitree-a1/}, 2024, accessed:
  2024-04-23.

\bibitem{drew2004cortical}
T.~Drew, S.~Prentice, and B.~Schepens, ``Cortical and brainstem control of
  locomotion,'' \emph{Progress in brain research}, vol. 143, pp. 251--261,
  2004.

\bibitem{saradjian2015sensory}
A.~Saradjian, ``Sensory modulation of movement, posture and locomotion,''
  \emph{Neurophysiologie Clinique/Clinical Neurophysiology}, vol.~45, no. 4-5,
  pp. 255--267, 2015.

\bibitem{ashtiani2021hybrid}
M.~S. Ashtiani, A.~Aghamaleki~Sarvestani, and A.~Badri-Spr{\"o}witz, ``Hybrid
  parallel compliance allows robots to operate with sensorimotor delays and low
  control frequencies,'' \emph{Frontiers in Robotics and AI}, vol.~8, p.
  645748, 2021.

\end{thebibliography}

\end{document}